\pdfoutput=1

\documentclass[11pt]{article}

\usepackage[preprint]{acl}

\usepackage{times}
\usepackage{latexsym}

\usepackage[T1]{fontenc}

\usepackage[utf8]{inputenc}

\usepackage{microtype}

\usepackage{inconsolata}

\usepackage{tabularx}

\usepackage{amssymb}
\usepackage{graphicx}
\graphicspath{ {./figs/} }

\usepackage{hyperref}
\usepackage{arydshln}

\usepackage{xcolor}

\newcommand\extrafootertext[1]{%
    \bgroup
    \renewcommand\thefootnote{\fnsymbol{footnote}}%
    \renewcommand\thempfootnote{\fnsymbol{mpfootnote}}%
    \footnotetext[0]{#1}%
    \egroup
}

\title{Accurate and Nuanced Open-QA Evaluation Through Textual Entailment}

\author{Peiran Yao \and Denilson Barbosa\\
Department of Computing Science\\
University of Alberta\\
{\tt \{peiran, denilson\}@ualberta.ca}}

\begin{document}
\maketitle
\begin{abstract}

Open-domain question answering (Open-QA) is a common task for evaluating large language models (LLMs).
However, current Open-QA evaluations are criticized for the ambiguity in questions and the lack of semantic understanding in evaluators.
Complex evaluators, powered by foundation models or LLMs and pertaining to semantic equivalence, still deviate from human judgments by a large margin.
We propose to study the entailment relations of answers to identify more informative and more general system answers, offering a much closer evaluation to human judgment on both NaturalQuestions and TriviaQA while being learning-free.
The entailment-based evaluation we propose allows the assignment of bonus or partial marks by quantifying the inference gap between answers, enabling a nuanced ranking of answer correctness that has higher AUC than current methods.

\end{abstract}

\section{Introduction}
\extrafootertext{\vspace{-3mm}\\To appear at ACL 2024 (Findings). Code and data are available at \url{https://github.com/U-Alberta/QA-partial-marks}.}

Open-domain question answering (Open-QA) is a long-established task requiring systems to generate precise answers to factual questions on any topic, from information in a large corpus of text \cite{voorhees-tice-2000-trec, zhang-et-al-2023-survey-efficient}.
A more restricted form of open-domain QA where answers are short is still regarded as challenging and as a reasonable test for the capabilities of recent large language models (LLMs) \cite{anil-et-al-2023-gemini,touvron-et-al-2023-llama2}, particularly when it comes to the assessment of LLM honesty \cite{yang-etal-2023-alignment}, calibration \cite{tian-etal-2023-just}. 
Open-QA benchmarks (\citealp{joshi-et-al-2017-triviaqa,kwiatkowski-et-al-2019-natural,lee-et-al-2019-latent}; \textit{inter alia}), consisting of pairs of curated questions and manually-annotated gold answers, have been under intense scrutiny because current automated evaluations have been found primitive, flawed, and insufficient to capture the true capabilities of Open-QA systems \cite{chen-et-al-2019-evaluating,boyd-graber-borschinger-2020-question,kamalloo-et-al-2023-evaluating,wang2023evaluating}.

\paragraph*{Open-QA Evaluators.}

Let $\mathcal{S}$ be the set of finite strings.
Given a question $q \in \mathcal{Q} \subset \mathcal{S}$, an Open-QA system generates a free-text \textit{system answer} $a \in \mathcal{A} \subset \mathcal{S}$, while reference \textit{gold answer}(s) $a^* \in \mathcal{A}^* \subset \mathcal{S}$ are provided by humans.
In the typical setting, an evaluator $f: \mathcal{Q} \times \mathcal{A} \times \mathcal{A}^* \mapsto \{0, 1\}$
compares the system answer $a$ with the gold answer(s) $a^*$ to provide an \textit{evaluator judgment} of whether the system correctly answered the question $q$.

While a wide variety of evaluators would be possible, current Open-QA benchmarks resort to fairly strict and primitive evaluators, which do a poor job with under-specified questions or when the system provides an answer that is either more general or more specific than the gold standard, and are believed to have hindered the understanding of LLM's ``emergent'' abilities \cite{schaeffer-etal-2023-emergent}.

\begin{figure*}[th]
    \centering
    \includegraphics[width=.96\linewidth]{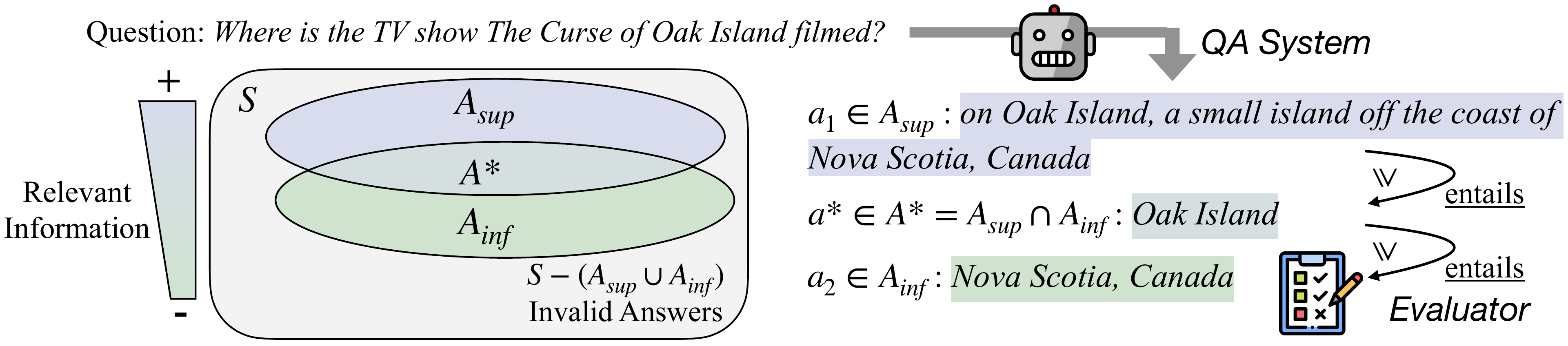}
    \caption{
        QA systems may generate a variety of correct answers that are neither exact matches nor semantic equivalents of the gold answer.
        Judging by the amount of information relevant to the gold answer that the system answers provide, we obtain a partial order of system answers with respect to the gold answer using textual entailment, and group answers into a hierarchy of subsets.
    }
    \label{fig:workflow}
\end{figure*}

\paragraph*{Ambiguity in Open-QA benchmarks.}
Questions from Open-QA benchmarks are often ambiguous and under-specified \cite{boyd-graber-borschinger-2020-question}, leading to multiple possible answers that are not always covered by the gold answers \cite{si-et-al-2021-whats}.
Figure~\ref{fig:workflow} presents an example from the NaturalQuestions benchmark \cite{kwiatkowski-et-al-2019-natural} where \textit{``Oak Island''} is the sole gold answer to the question \textit{``Where is the TV show The Curse of Oak Island filmed?''}.
However, due to the lack of specificity, a case can be made that \textit{more specific} answers such as \textit{``on Oak Island, a small island off the coast of Nova Scotia, Canada''}, or \textit{more general} answers such as \textit{``Nova Scotia, Canada''} should be accepted.
The former covers the gold answer and provides more details, while the latter has a lower level of specificity than the gold answer.
Exact word matching, a commonly used evaluator, fails with both answers. 
More advanced automated evaluations, including semantic similarity models and LLM in-context learning, are also shown to be incapable of capturing such intricacies and deviate from human judgment \cite{kamalloo-et-al-2023-evaluating,wang2023evaluating}.

\paragraph*{Our contributions.}
We study semantic relations between system answers and gold answers that consider whether system answers cover the gold answer while providing more details, or vice versa.
Naturally, textual entailment, the task of determining whether a piece of text entails or contradicts another, is a suitable and training-free tool for this categorization.
We propose to use textual entailment for the evaluation of Open-QA systems, and show that entailment-based evaluation metrics, even when used without finetuning, are consistent with human judgments and are more effective in capturing the true capabilities of a range of Open-QA systems when evaluating system answers on both NaturalQuestions (NQ) and TriviaQA (TQ; \citealp{joshi-et-al-2017-triviaqa}).
We also propose to use entailment-based evaluation metrics to assign bonus or partial marks to system answers by quantifying the inference gap between system answers and gold answers.
We argue that our metric offers a more informative and fairer alternative to current binary evaluation metrics.
Such a more accurate and nuanced QA evaluation scheme is valuable in solidifying the large body of concurrent studies that build on short-answer QA evaluations.

\section{Related Work}

Typical Open-QA evaluators rely on exact word match accuracy (lexical match), $F_1$ score over word matches (formally defined in \citet{bulian-et-al-2022-tomayto}), some semantic similarity model such as BERTScore \cite{zhang-et-al-2020-bertscore} or BLEURT \cite{sellam-das-parikh-2020-bleurt}, or zero-shot or in-context learning using an LLM \cite{chen-et-al-2023-llmzoo,kamalloo-et-al-2023-evaluating}.
These approaches have been under scrutiny, however.
\citet{bulian-et-al-2022-tomayto}, \citet{kamalloo-et-al-2023-evaluating} and \citet{wang2023evaluating} have looked at quantitatively assessing the correctness of automated evaluators by comparing the judgments they produce against \textit{gold judgments} produced by human experts.
They find that unsupervised automated evaluators, including those powered by pre-trained foundation models and LLMs, are not consistent with human judgments.

\citet{wang2023evaluating} released the \textsc{Evouna} dataset for the evaluation of automated evaluators, 
with 3,020 questions from NQ and 1,938 from TQ.
Questions are filtered to exclude those with outdated gold answers, and system answers generated by state-of-the-art Open-QA systems, namely DPR \cite{karpukhin-et-al-2020-dense} + FiD \cite{izacard-grave-2021-leveraging}, InstructGPT and ChatGPT \cite{ouyang-et-al-2022-instructgpt}, GPT-4 \cite{openai-2023-gpt4}, and BingChat,
are annotated with gold judgments.
An ideal automated evaluator should produce judgments that are consistent with human judgments, and thus, have higher $F_1$ scores and accuracies when evaluated against human judgments.
We base our work on \textsc{Evouna} and study the relation between gold answers and system answers, and subsequently derive an entailment-based evaluator $\hat{f}$ that is more consistent with human judgments. We also extend the range of $\hat{f}$ from a $\{0, 1\}$ binary prediction to $\mathbb{R}$ to achieve a more informative and fairer evaluation.

To overcome traditional metrics' lack of semantic understanding and, henceforth, underestimation of performance, recent efforts have focused on understanding the semantic equivalence of answers and developing evaluators accordingly \cite{si-et-al-2021-whats,bulian-et-al-2022-tomayto,kamalloo-et-al-2023-evaluating}.
We argue that the semantic equivalence is not the only relation between valid system answers and gold answers.
A valid answer can range from a vague, less informative one (\textit{e.g.}, a range, time period, or region) to a very specific and detailed answer (\textit{e.g.}, a precise number, time, or location).

\citet{bulian-et-al-2022-tomayto} proposed to accept all answers that contain at least all relevant content of the gold answer and no misleading content, while making no explicit distinction between semantic equivalence and entailment. 
We extend this idea and propose to assign a partial order to system answers with regard to how much relevant information the answer contains relative to the gold answer.

\section{The Answer Hierarchy}
\label{sec:hierarchy}

\begin{table*}[t]
    \centering
    \resizebox{0.89\textwidth}{!}{
    \begin{tabular}{cllllllllll}
    \hline
                  & \multicolumn{2}{c}{\textbf{DPR-FiD}}                & \multicolumn{2}{c}{\textbf{InstructGPT}}            & \multicolumn{2}{c}{\textbf{ChatGPT}}                & \multicolumn{2}{c}{\textbf{GPT-4}}                  & \multicolumn{2}{c}{\textbf{BingChat}}               \\
                  Evaluator & \multicolumn{1}{c}{$F_1$} & \multicolumn{1}{c}{Acc} & \multicolumn{1}{c}{$F_1$} & \multicolumn{1}{c}{Acc} & \multicolumn{1}{c}{$F_1$} & \multicolumn{1}{c}{Acc} & \multicolumn{1}{c}{$F_1$} & \multicolumn{1}{c}{Acc} & \multicolumn{1}{c}{$F_1$} & \multicolumn{1}{c}{Acc} \\ \hline
    Lexical Match$^\dagger$ & 92.0                      & 89.7                    & 86.9                      & 84.8                    & 85.0                      & 80.3                    & 87.6                      & 82.5                    & 87.8                      & 82.3                    \\
    BERTScore$^\dagger$     & 83.5                      & 75.1                    & 77.6                      & 69.5                    & 81.2                      & 72.8                    & 84.3                      & 76.0                    & 77.5                      & 67.5                    \\
    GPT-3.5$^\dagger$       & \textbf{95.3}             & \textbf{93.6}           & 87.2                      & 84.1                    & 86.9                      & 82.2                    & 86.8                      & 80.9                    & 77.3                      & 69.5                    \\
    \textbf{Entailment}    & 94.8                      & 92.5                    & \textbf{92.7}             & \textbf{90.2}           & \textbf{92.6}             & \textbf{88.9}           & \textbf{93.8}             & \textbf{90.1}           & \textbf{92.6}             & \textbf{88.1}           \\\hdashline[1pt/1pt]
    Entailment (small) & 91.5 & 88.5&88.0&85.4&87.7&83.2&89.9&85.0&87.8&82.0\\
    GPT-3.5 (best prompting) & 95.5 & 93.9 & 88.3 & 84.5 & 89.4 & 84.5 & 91.2 & 86.0 & 87.1 & 80.4\\ 
    Another Human$^\dagger$ & 97.4                      & 96.3                    & 97.8                      & 96.8                    & 96.5                      & 95.6                    & 97.9                      & 96.6                    & 97.2                      & 95.5                    \\ \hline
    \multicolumn{11}{c}{on \textsc{Evouna}-NaturalQuestions}                                                                                                                                                                                                                                             \\ \hline
    Lexical Match$^\dagger$ & 91.8                      & 94.7                    & 94.8                      & 92.3                    & 95.2                      & 92.3                    & 94.8                      & 91.1                    & 94.1                      & 89.8                    \\
    BERTScore$^\dagger$     & 75.1                      & 65.5                    & 84.1                      & 75.7                    & 88.4                      & 80.8                    & 90.5                      & 93.5                    & 88.3                      & 80.4                    \\
    GPT-3.5$^\dagger$       & \textbf{97.3}             & \textbf{95.7}           & 94.2                      & 91.2                    & 95.5                      & 92.5                    & 95.7                      & 92.4                    & 88.2                      & 80.9                    \\
    \textbf{Entailment}    & 96.8                      & 94.7                    & \textbf{96.6}             & \textbf{94.2}           & \textbf{96.6}             & \textbf{94.2}           & \textbf{97.4}             & \textbf{95.3}           & \textbf{95.9}             & \textbf{92.5}           \\\hdashline[1pt/1pt]
    Another Human$^\dagger$ & 100                       & 100                     & 99.6                      & 99.4                    & 99.2                      & 98.8                    & 99.2                      & 99.8                    & 99.9                      & 95.5                    \\ \hline
    \multicolumn{11}{c}{on \textsc{Evouna}-TriviaQA}                                                                                                                                                                                                                                                     \\ \hline
    \end{tabular}
    } %

    \caption{\label{tab:eval-eval}
    Using human judgments as the gold standard, entailment-based evaluation of Open-QA systems on both NQ and TQ yields higher $F_1$ scores and accuracies than lexical match, BERTScore, and GPT-3.5 when evaluating the judgments against gold judgments in \textsc{Evouna}.
    Metrics chosen following \citet{wang2023evaluating}.
    Higher scores and accuracies indicate that evaluator judgments are more consistent with human judgment.
    Judgments from another human are included as a reference of the upper bounds induced by ambiguity and inconsistencies in creating the gold judgments.
    Top performing evaluators are in bold.
    $\dagger$: scores reported by \citet{wang2023evaluating}.
    }
    \end{table*}

Let $A^*$ denote the set of gold answers for an Open-QA benchmark. We define two other sets of answers: $A_{\mathit{sup}}$ is the set of \textit{superior} answers that provide more information than what is in the gold standard, and $A_{\mathit{inf}}$ is the set of \textit{inferior} answers that only address the question partially. 
Given a system answer $a$ and the corresponding gold answer $a^*$, we say that $a \in A_{sup}$ if and only if $a$ entails $a^*$ within the context of the question. (This naturally extends to the case where multiple gold answers are given).
Similarly, we say that $a \in A_{inf}$ if and only if it is entailed by $a^*$.
Finally, an answer would be incorrect if and only if $a$ and $a^*$ are not entailed by each other; in this case $a \in S-(A_{sup}\cup A_{inf})$.
As a special case, it follows that answer $a$ is equivalent to the gold answer $a^*$ \textit{iff.} $a\in A_{sup} \cap A_{inf}$.

Before textual entailment, question-answer pairs are rewritten as declarative statements (introduced by \citet{demszky-etal-2018-qa2d} as QA2D) using GPT-3.5, as the question and context are important for assessing answers \cite{kamalloo-et-al-2023-evaluating}. For example, the gold answer in Figure~\ref{fig:workflow} is converted into the statement ``\textit{The TV show The Curse of Oak Island is filmed on Oak Island}''.
Inspired by LLM's strong performance in natural language inference tasks \cite{qin-et-al-2023-chatgpt}, we use GPT-3.5 to conduct textual entailment tests.
In the finalized approach, two steps, converting a question-answer pair to a declarative statement, and performing textual entailment test on declarative statements, are perform by a LLM in a few-shot manner.
We validate that for both steps GPT-3.5 as the LLM achieves high statistical reliability (Appendix~\ref{app:reliability}) in terms of the agreement across different seeds, and high validity (Appendix~\ref{app:validity}) in terms of the alignment with human labels.
Implementation details and a worked example are provided in Appendix~\ref{app:settings}.

\begin{table}[t]
    \centering
    \begin{tabular}{lll}
    \hline
                                 \textbf{Method} & \textbf{F}$_1$ & \textbf{Acc}  \\ \hline
    Llama-2 (SFT)       & 94.6  & 92.3 \\
    Llama-2 + NLI (SFT)     & 94.8     & 92.6    \\
    CVI                          & 84.7  & 73.5 \\
    \textbf{Entailment (0-shot)} & 93.5  & 90.2 \\ \hline
    \end{tabular}
    \caption{\label{tab:eval-sft}Without doing supervised finetuning (SFT), entailment-based evaluation yields comparable performance to data-driven approaches like finetuned Llama-2-7B and CVI when evaluating system answers on NQ.
    }

\end{table}

\paragraph*{System answers deserve partial credits and bonus credits.}
\label{sec:stat-test}
The above-described entailment-based evaluator reveals that $A_{sup} \oplus A_{inf}$ (the disjoint union) represents a considerable amount of valid system answers that would otherwise be disregarded (Table~\ref{tab:nouva-nq-hierarchy}, \ref{tab:nouva-tq-hierarchy}).
Assuming the hierarchy holds, we would see a $10.1\%$ and a $6.5\%$ increase in accuracy for NQ and TQ, respectively, which account for the reported underestimation of QA performance \cite{bulian-et-al-2022-tomayto,wang2023evaluating}.
We validate the hierarchy by demonstrating that higher positions correspond to better answers judged by humans. 
This is supported by one-tailed Fisher's exact tests, all yielding significant results with $p < 0.01$ with the exception of DPR-FiD and $A_{sup}-A^\star$ in TQ. Details of the statistical tests can be found in Appendix~\ref{app:interset}.

\paragraph{The answer hierarchy is a superior automated evaluator.}
Treating $A_{sup} \cup A_{inf}$ as correct answers\footnote{Alternative choices discussed in Appendix \ref{app:hierarchy}.} bring the evaluation results of various systems closer to human performance, as shown in Table~\ref{tab:eval-eval}, and confirms the observation that current evaluators\footnote{Details about baselines in Appendix~\ref{app:eval-baseline}.} unfairly misrepresent the capabilities of those systems.
However, unlike previous studies which resorted to manual inspections of answers, our evaluator allows the same observation in an automated way. Moreover, our evaluator can be used for any benchmark.
The system where entailment does not improve the performance is DPR-FiD, which is an extractive model that outputs a span of text that requires less semantic understanding to evaluate than complete sentences. Nevertheless, the entailment evaluator assessed that system very closely to the numbers reported in the literature.

\paragraph{Although learning-free, entailment is comparable to finetuned evaluators.}

\citet{bulian-et-al-2022-tomayto} and \citet{kamalloo-et-al-2023-evaluating} advocate for learned evaluators to close the gap between automated and human evaluation.
For comparison, we partition \textsc{Evouna}-NQ by questions into 50:50 train/test splits, and finetune a Llama-2 \cite{touvron-et-al-2023-llama2} model which only performs slightly better than our method.
We show that explicitly including entailment as a feature improves the finetuned model (\textit{+NLI} in Table~\ref{tab:eval-sft}).
Moreover, we finetune another Llama-2 model with the same training data, but with system answers as contexts to predict gold answers, in order to use conditional $\mathcal{V}$-information (CVI; \citealp{hewitt-et-al-2021-cvi,chen-etal-2023-rev}) as another evaluator that builds on usable information from the system answers. The results are shown in Table~\ref{tab:eval-sft}.
Entailment also yields better results than in-context learning with four examples (\textit{best prompting} in Table~\ref{tab:eval-eval}). Details of the finetuning process and the learned baselines are in Appendix~\ref{app:eval-sft}.

\paragraph{Out-of-the-box entailment outperforms prompt engineering.}
Since our method is implemented solely with GPT-3.5, it can be seen as a prompt/flow-engineering method that outperforms the best prompt engineering technique among those \citet{kamalloo-et-al-2023-evaluating} extensively explored (\textit{best prompting} in Table~\ref{tab:eval-eval}), such as Chain-of-Thought \cite{wei-et-al-2022-cot} and in-context learning.
Meanwhile, the pre-processing and entailment tests can be implemented independent of LLMs, for example using DeBERTa \cite{he-gao-chen-2023-debertav3} as the NLI model and Llama-2-7B as the question to statement conversion model (\textit{small} in Table~\ref{tab:eval-eval}), while still achieving improved results.

\section{Towards Partial Marks}

\citet{bulian-et-al-2022-tomayto} demonstrated by examples that the seemingly continuous $F_1$ score is not indicative of how close the system answer is to the gold answer.
Going beyond directly using the classification probability from the NLI model \cite{chen-etal-2021-nli-models}, we hypothesize that measuring the inference gap, \textit{i.e.} how many steps, assumptions, and additional pieces of information are needed to derive a system answer $a$ from a gold answer $a^*$, can be used to assign partial marks in a way that reflects semantic closeness.
Inspired by Chain-of-Thought prompting, explainable natural language inference \cite{camburu-etal-2018-esnli}, and LLM-based decompositions of implicit content \cite{hoyle-etal-2023-natural}, we propose to use LLM (GPT-3.5 in our experiments) to explain step-by-step the inference process behind how $a^*$ entails $a$, along with assumptions and additional knowledge required (\textit{``Inference''}).
Based on the explanation, a score of inference difficulty is directly produced by the LLM (\textit{``LLM Score''}), or the number of steps is counted (\textit{``\#Steps''}) as partial marks.
Details of the scoring schemes, worked examples, and alternatives are discussed in Appendix~\ref{app:partial}.

We have examined the inter-set ranks in the answer hierarchy in \S\ref{sec:stat-test}. When it comes to the intra-set ranking of partial answers in $A_{inf}-A_{sup}$, Brunner-Munzel tests \cite{brunner-munzel-2000-nonparametric} show that both LLM Score and \#Steps, as well as other baselines in Table~\ref{tab:partial}, assign higher scores to human-accepted answers ($p < 0.001$).
Quantitatively, Table~\ref{tab:partial} shows that the inference-process-based scores have higher AUROC on \textsc{Evouna}-NQ than $F_1$ score or using GPT-3.5 to directly assess system answers on a 5-point scale (``\textit{LLM Score}''), indicating that partial marks assigned by our method are suitable for capturing the nuanced goodness differences between system answers.

\begin{table}[t!]
    \centering
    
    \begin{tabular}{lr}
    \hline
    \textbf{Method}                & \textbf{AUC} \\ \hline
    Inference + LLM Score & 0.91  \\
    Inference + \#Steps    & 0.91  \\
    LLM Score             & 0.88  \\
    $F_1$ Score           & 0.78  \\ \hline
    \end{tabular}
    \caption{Using LLM to explain the inference process behind how gold answers entail the system answers leads to higher AUROC in predicting human judgements on NQ, making it a good candidate for partial marks.}
    \label{tab:partial}
\end{table}

\section{Conclusion}
In theory, textual entailment is considered AI-Complete \cite{dagan-2009-nli-complete} - an embodiment of general AI that solves all AI tasks, Open-QA evaluation included.
In practice, we showed that state-of-the-art textual entailment provides a simple yet powerful replacement for Open-QA evaluation, and it offers the prospect of soft and partial marks.

\section*{Limitations}
Our current work studies the potential of textual entailment as a fairer and finer-grained replacement for Open-QA evaluation. We only explored using the method for benchmarking QA systems. However, it remains a highly interesting topic to investigate how it can be used as a softer signal for training QA systems with the potential of improvements, given the success of smoothed labels \cite{hinton-2015-kd,szegedy-etal-2016-smoothing}.
We only studied QA benchmarks consisting of mostly factoid questions and relatively short and simple answers.
For QA tasks that require more complex, and potentially multi-passage multi-facet answers, it is unclear how well the original entailment method can be directly applied.
Future work is required to investigate the entailment relations and the matching between multiple units of meaning, such as in \citet{10.1162/tacl_a_00453}, to extend our work to more complex QA tasks. 

\section*{Acknowledgements}
We acknowledge the support of the Natural Sciences and Engineering Research Council of Canada (NSERC). This work is also supported in part by a gift from Scotiabank. Icons in Figure~\ref{fig:workflow} are designed by \href{https://openmoji.org/}{OpenMoji} under CC BY-SA 4.0 license, and by Flaticon-Freepik.

\bibliography{main}

\begin{thebibliography}{42}
\expandafter\ifx\csname natexlab\endcsname\relax\def\natexlab#1{#1}\fi

\bibitem[{Anil et~al.(2023)Anil, Borgeaud, Wu, Alayrac, Yu, Soricut, Schalkwyk,
  Dai, Hauth, Millican, Silver, Petrov, Johnson, Antonoglou, Schrittwieser,
  Glaese, Chen, Pitler, Lillicrap, Lazaridou, Firat, Molloy, Isard, Barham,
  Hennigan, Lee, Viola, Reynolds, Xu, Doherty, Collins, Meyer, Rutherford,
  Moreira, Ayoub, Goel, Tucker, Piqueras, Krikun, Barr, Savinov, Danihelka,
  Roelofs, White, Andreassen, von Glehn, Yagati, Kazemi, Gonzalez, Khalman,
  Sygnowski, and et~al.}]{anil-et-al-2023-gemini}
Rohan Anil, Sebastian Borgeaud, Yonghui Wu, Jean{-}Baptiste Alayrac, Jiahui Yu,
  Radu Soricut, Johan Schalkwyk, Andrew~M. Dai, Anja Hauth, Katie Millican,
  David Silver, Slav Petrov, Melvin Johnson, Ioannis Antonoglou, Julian
  Schrittwieser, Amelia Glaese, Jilin Chen, Emily Pitler, Timothy~P. Lillicrap,
  Angeliki Lazaridou, Orhan Firat, James Molloy, Michael Isard, Paul~Ronald
  Barham, Tom Hennigan, Benjamin Lee, Fabio Viola, Malcolm Reynolds, Yuanzhong
  Xu, Ryan Doherty, Eli Collins, Clemens Meyer, Eliza Rutherford, Erica
  Moreira, Kareem Ayoub, Megha Goel, George Tucker, Enrique Piqueras, Maxim
  Krikun, Iain Barr, Nikolay Savinov, Ivo Danihelka, Becca Roelofs,
  Ana{\"{\i}}s White, Anders Andreassen, Tamara von Glehn, Lakshman Yagati,
  Mehran Kazemi, Lucas Gonzalez, Misha Khalman, Jakub Sygnowski, and et~al.
  2023.
\newblock \href {https://doi.org/10.48550/ARXIV.2312.11805} {Gemini: {A} family
  of highly capable multimodal models}.
\newblock \emph{CoRR}, abs/2312.11805.

\bibitem[{Boyd-Graber and
  B{\"o}rschinger(2020)}]{boyd-graber-borschinger-2020-question}
Jordan Boyd-Graber and Benjamin B{\"o}rschinger. 2020.
\newblock \href {https://doi.org/10.18653/v1/2020.acl-main.662} {What question
  answering can learn from trivia nerds}.
\newblock In \emph{Proceedings of the 58th Annual Meeting of the Association
  for Computational Linguistics}, pages 7422--7435, Online. Association for
  Computational Linguistics.

\bibitem[{Brunner and Munzel(2000)}]{brunner-munzel-2000-nonparametric}
Edgar Brunner and Ullrich Munzel. 2000.
\newblock \href
  {https://doi.org/https://doi.org/10.1002/(SICI)1521-4036(200001)42:1<17::AID-BIMJ17>3.0.CO;2-U}
  {The nonparametric behrens-fisher problem: Asymptotic theory and a
  small-sample approximation}.
\newblock \emph{Biometrical Journal}, 42(1):17--25.

\bibitem[{Bulian et~al.(2022)Bulian, Buck, Gajewski, B{\"o}rschinger, and
  Schuster}]{bulian-et-al-2022-tomayto}
Jannis Bulian, Christian Buck, Wojciech Gajewski, Benjamin B{\"o}rschinger, and
  Tal Schuster. 2022.
\newblock \href {https://doi.org/10.18653/v1/2022.emnlp-main.20} {Tomayto,
  tomahto. beyond token-level answer equivalence for question answering
  evaluation}.
\newblock In \emph{Proceedings of the 2022 Conference on Empirical Methods in
  Natural Language Processing}, pages 291--305, Abu Dhabi, United Arab
  Emirates. Association for Computational Linguistics.

\bibitem[{Camburu et~al.(2018)Camburu, Rockt\"{a}schel, Lukasiewicz, and
  Blunsom}]{camburu-etal-2018-esnli}
Oana-Maria Camburu, Tim Rockt\"{a}schel, Thomas Lukasiewicz, and Phil Blunsom.
  2018.
\newblock \href
  {https://proceedings.neurips.cc/paper_files/paper/2018/file/4c7a167bb329bd92580a99ce422d6fa6-Paper.pdf}
  {e-snli: Natural language inference with natural language explanations}.
\newblock In \emph{Advances in Neural Information Processing Systems},
  volume~31. Curran Associates, Inc.

\bibitem[{Chen et~al.(2019)Chen, Stanovsky, Singh, and
  Gardner}]{chen-et-al-2019-evaluating}
Anthony Chen, Gabriel Stanovsky, Sameer Singh, and Matt Gardner. 2019.
\newblock \href {https://doi.org/10.18653/v1/D19-5817} {Evaluating question
  answering evaluation}.
\newblock In \emph{Proceedings of the 2nd Workshop on Machine Reading for
  Question Answering}, pages 119--124, Hong Kong, China. Association for
  Computational Linguistics.

\bibitem[{Chen et~al.(2023{\natexlab{a}})Chen, Brahman, Ren, Ji, Choi, and
  Swayamdipta}]{chen-etal-2023-rev}
Hanjie Chen, Faeze Brahman, Xiang Ren, Yangfeng Ji, Yejin Choi, and Swabha
  Swayamdipta. 2023{\natexlab{a}}.
\newblock \href {https://doi.org/10.18653/v1/2023.acl-long.112} {{REV}:
  Information-theoretic evaluation of free-text rationales}.
\newblock In \emph{Proceedings of the 61st Annual Meeting of the Association
  for Computational Linguistics (Volume 1: Long Papers)}, pages 2007--2030,
  Toronto, Canada. Association for Computational Linguistics.

\bibitem[{Chen et~al.(2021)Chen, Choi, and Durrett}]{chen-etal-2021-nli-models}
Jifan Chen, Eunsol Choi, and Greg Durrett. 2021.
\newblock \href {https://doi.org/10.18653/v1/2021.findings-emnlp.324} {Can
  {NLI} models verify {QA} systems{'} predictions?}
\newblock In \emph{Findings of the Association for Computational Linguistics:
  EMNLP 2021}, pages 3841--3854, Punta Cana, Dominican Republic. Association
  for Computational Linguistics.

\bibitem[{Chen et~al.(2023{\natexlab{b}})Chen, Jiang, Chen, Wang, Yu, Chen,
  Zhang, Liang, Zhang, Zhang, Li, Wan, Wang, and Li}]{chen-et-al-2023-llmzoo}
Zhihong Chen, Feng Jiang, Junying Chen, Tiannan Wang, Fei Yu, Guiming Chen,
  Hongbo Zhang, Juhao Liang, Chen Zhang, Zhiyi Zhang, Jianquan Li, Xiang Wan,
  Benyou Wang, and Haizhou Li. 2023{\natexlab{b}}.
\newblock \href {https://doi.org/10.48550/ARXIV.2304.10453} {Phoenix:
  Democratizing {ChatGPT} across languages}.
\newblock \emph{CoRR}, abs/2304.10453.

\bibitem[{Dagan et~al.(2009)Dagan, Dolan, Magnini, and
  Roth}]{dagan-2009-nli-complete}
Ido Dagan, Bill Dolan, Bernado Magnini, and Dan Roth. 2009.
\newblock \href {https://doi.org/10.1017/S1351324909990209} {Recognizing
  textual entailment: Rational, evaluation and approaches}.
\newblock \emph{Natural Language Engineering}, 15(4):i–xvii.

\bibitem[{Demszky et~al.(2018)Demszky, Guu, and Liang}]{demszky-etal-2018-qa2d}
Dorottya Demszky, Kelvin Guu, and Percy Liang. 2018.
\newblock \href {http://arxiv.org/abs/1809.02922} {Transforming question
  answering datasets into natural language inference datasets}.
\newblock \emph{CoRR}, abs/1809.02922.

\bibitem[{He et~al.(2023)He, Gao, and Chen}]{he-gao-chen-2023-debertav3}
Pengcheng He, Jianfeng Gao, and Weizhu Chen. 2023.
\newblock \href {https://openreview.net/pdf?id=sE7-XhLxHA} {Debertav3:
  Improving deberta using electra-style pre-training with gradient-disentangled
  embedding sharing}.
\newblock In \emph{The Eleventh International Conference on Learning
  Representations, {ICLR} 2023, Kigali, Rwanda, May 1-5, 2023}. OpenReview.net.

\bibitem[{Hewitt et~al.(2021)Hewitt, Ethayarajh, Liang, and
  Manning}]{hewitt-et-al-2021-cvi}
John Hewitt, Kawin Ethayarajh, Percy Liang, and Christopher~D. Manning. 2021.
\newblock \href {https://doi.org/10.18653/V1/2021.EMNLP-MAIN.122} {Conditional
  probing: measuring usable information beyond a baseline}.
\newblock In \emph{Proceedings of the 2021 Conference on Empirical Methods in
  Natural Language Processing, {EMNLP} 2021, Virtual Event / Punta Cana,
  Dominican Republic, 7-11 November, 2021}, pages 1626--1639. Association for
  Computational Linguistics.

\bibitem[{Hinton et~al.(2015)Hinton, Vinyals, and Dean}]{hinton-2015-kd}
Geoffrey Hinton, Oriol Vinyals, and Jeffrey Dean. 2015.
\newblock \href {http://arxiv.org/abs/1503.02531} {Distilling the knowledge in
  a neural network}.
\newblock In \emph{NIPS Deep Learning and Representation Learning Workshop}.

\bibitem[{Hoyle et~al.(2023)Hoyle, Sarkar, Goel, and
  Resnik}]{hoyle-etal-2023-natural}
Alexander Hoyle, Rupak Sarkar, Pranav Goel, and Philip Resnik. 2023.
\newblock \href {https://doi.org/10.18653/v1/2023.emnlp-main.815} {Natural
  language decompositions of implicit content enable better text
  representations}.
\newblock In \emph{Proceedings of the 2023 Conference on Empirical Methods in
  Natural Language Processing}, pages 13188--13214, Singapore. Association for
  Computational Linguistics.

\bibitem[{Izacard and Grave(2021)}]{izacard-grave-2021-leveraging}
Gautier Izacard and Edouard Grave. 2021.
\newblock \href {https://doi.org/10.18653/v1/2021.eacl-main.74} {Leveraging
  passage retrieval with generative models for open domain question answering}.
\newblock In \emph{Proceedings of the 16th Conference of the European Chapter
  of the Association for Computational Linguistics: Main Volume}, pages
  874--880, Online. Association for Computational Linguistics.

\bibitem[{Joshi et~al.(2017)Joshi, Choi, Weld, and
  Zettlemoyer}]{joshi-et-al-2017-triviaqa}
Mandar Joshi, Eunsol Choi, Daniel Weld, and Luke Zettlemoyer. 2017.
\newblock \href {https://doi.org/10.18653/v1/P17-1147} {{T}rivia{QA}: A large
  scale distantly supervised challenge dataset for reading comprehension}.
\newblock In \emph{Proceedings of the 55th Annual Meeting of the Association
  for Computational Linguistics (Volume 1: Long Papers)}, pages 1601--1611,
  Vancouver, Canada. Association for Computational Linguistics.

\bibitem[{Kamalloo et~al.(2023)Kamalloo, Dziri, Clarke, and
  Rafiei}]{kamalloo-et-al-2023-evaluating}
Ehsan Kamalloo, Nouha Dziri, Charles Clarke, and Davood Rafiei. 2023.
\newblock \href {https://doi.org/10.18653/v1/2023.acl-long.307} {Evaluating
  open-domain question answering in the era of large language models}.
\newblock In \emph{Proceedings of the 61st Annual Meeting of the Association
  for Computational Linguistics (Volume 1: Long Papers)}, pages 5591--5606,
  Toronto, Canada. Association for Computational Linguistics.

\bibitem[{Karpukhin et~al.(2020)Karpukhin, Oguz, Min, Lewis, Wu, Edunov, Chen,
  and Yih}]{karpukhin-et-al-2020-dense}
Vladimir Karpukhin, Barlas Oguz, Sewon Min, Patrick Lewis, Ledell Wu, Sergey
  Edunov, Danqi Chen, and Wen-tau Yih. 2020.
\newblock \href {https://doi.org/10.18653/v1/2020.emnlp-main.550} {Dense
  passage retrieval for open-domain question answering}.
\newblock In \emph{Proceedings of the 2020 Conference on Empirical Methods in
  Natural Language Processing (EMNLP)}, pages 6769--6781, Online. Association
  for Computational Linguistics.

\bibitem[{Kwiatkowski et~al.(2019)Kwiatkowski, Palomaki, Redfield, Collins,
  Parikh, Alberti, Epstein, Polosukhin, Devlin, Lee, Toutanova, Jones, Kelcey,
  Chang, Dai, Uszkoreit, Le, and Petrov}]{kwiatkowski-et-al-2019-natural}
Tom Kwiatkowski, Jennimaria Palomaki, Olivia Redfield, Michael Collins, Ankur
  Parikh, Chris Alberti, Danielle Epstein, Illia Polosukhin, Jacob Devlin,
  Kenton Lee, Kristina Toutanova, Llion Jones, Matthew Kelcey, Ming-Wei Chang,
  Andrew~M. Dai, Jakob Uszkoreit, Quoc Le, and Slav Petrov. 2019.
\newblock \href {https://doi.org/10.1162/tacl_a_00276} {Natural questions: A
  benchmark for question answering research}.
\newblock \emph{Transactions of the Association for Computational Linguistics},
  7:452--466.

\bibitem[{Laban et~al.(2022)Laban, Schnabel, Bennett, and
  Hearst}]{10.1162/tacl_a_00453}
Philippe Laban, Tobias Schnabel, Paul~N. Bennett, and Marti~A. Hearst. 2022.
\newblock \href {https://doi.org/10.1162/tacl_a_00453} {{SummaC: Re-Visiting
  NLI-based Models for Inconsistency Detection in Summarization}}.
\newblock \emph{Transactions of the Association for Computational Linguistics},
  10:163--177.

\bibitem[{Landis and Koch(1977)}]{landis-koch-1977-kappa}
J.~Richard Landis and Gary~G. Koch. 1977.
\newblock \href {http://www.jstor.org/stable/2529310} {The measurement of
  observer agreement for categorical data}.
\newblock \emph{Biometrics}, 33(1):159--174.

\bibitem[{Lee et~al.(2019)Lee, Chang, and Toutanova}]{lee-et-al-2019-latent}
Kenton Lee, Ming-Wei Chang, and Kristina Toutanova. 2019.
\newblock \href {https://doi.org/10.18653/v1/P19-1612} {Latent retrieval for
  weakly supervised open domain question answering}.
\newblock In \emph{Proceedings of the 57th Annual Meeting of the Association
  for Computational Linguistics}, pages 6086--6096, Florence, Italy.
  Association for Computational Linguistics.

\bibitem[{Mangrulkar et~al.(2022)Mangrulkar, Gugger, Debut, Belkada, Paul, and
  Bossan}]{peft}
Sourab Mangrulkar, Sylvain Gugger, Lysandre Debut, Younes Belkada, Sayak Paul,
  and Benjamin Bossan. 2022.
\newblock Peft: State-of-the-art parameter-efficient fine-tuning methods.
\newblock \url{https://github.com/huggingface/peft}.

\bibitem[{OpenAI(2023)}]{openai-2023-gpt4}
OpenAI. 2023.
\newblock \href {https://doi.org/10.48550/ARXIV.2303.08774} {{GPT-4} technical
  report}.
\newblock \emph{CoRR}, abs/2303.08774.

\bibitem[{Ouyang et~al.(2022)Ouyang, Wu, Jiang, Almeida, Wainwright, Mishkin,
  Zhang, Agarwal, Slama, Ray, Schulman, Hilton, Kelton, Miller, Simens, Askell,
  Welinder, Christiano, Leike, and Lowe}]{ouyang-et-al-2022-instructgpt}
Long Ouyang, Jeffrey Wu, Xu~Jiang, Diogo Almeida, Carroll Wainwright, Pamela
  Mishkin, Chong Zhang, Sandhini Agarwal, Katarina Slama, Alex Ray, John
  Schulman, Jacob Hilton, Fraser Kelton, Luke Miller, Maddie Simens, Amanda
  Askell, Peter Welinder, Paul~F Christiano, Jan Leike, and Ryan Lowe. 2022.
\newblock \href
  {https://proceedings.neurips.cc/paper_files/paper/2022/file/b1efde53be364a73914f58805a001731-Paper-Conference.pdf}
  {Training language models to follow instructions with human feedback}.
\newblock In \emph{Advances in Neural Information Processing Systems},
  volume~35, pages 27730--27744. Curran Associates, Inc.

\bibitem[{Qin et~al.(2023)Qin, Zhang, Zhang, Chen, Yasunaga, and
  Yang}]{qin-et-al-2023-chatgpt}
Chengwei Qin, Aston Zhang, Zhuosheng Zhang, Jiaao Chen, Michihiro Yasunaga, and
  Diyi Yang. 2023.
\newblock \href {https://doi.org/10.18653/v1/2023.emnlp-main.85} {Is
  {C}hat{GPT} a general-purpose natural language processing task solver?}
\newblock In \emph{Proceedings of the 2023 Conference on Empirical Methods in
  Natural Language Processing}, pages 1339--1384, Singapore. Association for
  Computational Linguistics.

\bibitem[{Reimers and Gurevych(2019)}]{reimers-gurevych-2019-sbert}
Nils Reimers and Iryna Gurevych. 2019.
\newblock \href {https://doi.org/10.18653/v1/D19-1410} {Sentence-{BERT}:
  Sentence embeddings using {S}iamese {BERT}-networks}.
\newblock In \emph{Proceedings of the 2019 Conference on Empirical Methods in
  Natural Language Processing and the 9th International Joint Conference on
  Natural Language Processing (EMNLP-IJCNLP)}, pages 3982--3992, Hong Kong,
  China. Association for Computational Linguistics.

\bibitem[{Schaeffer et~al.(2023)Schaeffer, Miranda, and
  Koyejo}]{schaeffer-etal-2023-emergent}
Rylan Schaeffer, Brando Miranda, and Sanmi Koyejo. 2023.
\newblock \href
  {https://proceedings.neurips.cc/paper_files/paper/2023/file/adc98a266f45005c403b8311ca7e8bd7-Paper-Conference.pdf}
  {Are emergent abilities of large language models a mirage?}
\newblock In \emph{Advances in Neural Information Processing Systems},
  volume~36, pages 55565--55581. Curran Associates, Inc.

\bibitem[{Sellam et~al.(2020)Sellam, Das, and
  Parikh}]{sellam-das-parikh-2020-bleurt}
Thibault Sellam, Dipanjan Das, and Ankur~P. Parikh. 2020.
\newblock \href {https://doi.org/10.18653/V1/2020.ACL-MAIN.704} {{BLEURT:}
  learning robust metrics for text generation}.
\newblock In \emph{Proceedings of the 58th Annual Meeting of the Association
  for Computational Linguistics, {ACL} 2020, Online, July 5-10, 2020}, pages
  7881--7892. Association for Computational Linguistics.

\bibitem[{Si et~al.(2021)Si, Zhao, and Boyd-Graber}]{si-et-al-2021-whats}
Chenglei Si, Chen Zhao, and Jordan Boyd-Graber. 2021.
\newblock \href {https://doi.org/10.18653/v1/2021.emnlp-main.757} {What{'}s in
  a name? answer equivalence for open-domain question answering}.
\newblock In \emph{Proceedings of the 2021 Conference on Empirical Methods in
  Natural Language Processing}, pages 9623--9629, Online and Punta Cana,
  Dominican Republic. Association for Computational Linguistics.

\bibitem[{Szegedy et~al.(2016)Szegedy, Vanhoucke, Ioffe, Shlens, and
  Wojna}]{szegedy-etal-2016-smoothing}
Christian Szegedy, Vincent Vanhoucke, Sergey Ioffe, Jonathon Shlens, and
  Zbigniew Wojna. 2016.
\newblock \href {https://doi.org/10.1109/CVPR.2016.308} {Rethinking the
  inception architecture for computer vision}.
\newblock In \emph{2016 {IEEE} Conference on Computer Vision and Pattern
  Recognition, {CVPR} 2016, Las Vegas, NV, USA, June 27-30, 2016}, pages
  2818--2826. {IEEE} Computer Society.

\bibitem[{Tian et~al.(2023)Tian, Mitchell, Zhou, Sharma, Rafailov, Yao, Finn,
  and Manning}]{tian-etal-2023-just}
Katherine Tian, Eric Mitchell, Allan Zhou, Archit Sharma, Rafael Rafailov,
  Huaxiu Yao, Chelsea Finn, and Christopher Manning. 2023.
\newblock \href {https://doi.org/10.18653/v1/2023.emnlp-main.330} {Just ask for
  calibration: Strategies for eliciting calibrated confidence scores from
  language models fine-tuned with human feedback}.
\newblock In \emph{Proceedings of the 2023 Conference on Empirical Methods in
  Natural Language Processing}, pages 5433--5442, Singapore. Association for
  Computational Linguistics.

\bibitem[{Touvron et~al.(2023)Touvron, Martin, Stone, Albert, Almahairi,
  Babaei, Bashlykov, Batra, Bhargava, Bhosale, Bikel, Blecher, Canton{-}Ferrer,
  Chen, Cucurull, Esiobu, Fernandes, Fu, Fu, Fuller, Gao, Goswami, Goyal,
  Hartshorn, Hosseini, Hou, Inan, Kardas, Kerkez, Khabsa, Kloumann, Korenev,
  Koura, Lachaux, Lavril, Lee, Liskovich, Lu, Mao, Martinet, Mihaylov, Mishra,
  Molybog, Nie, Poulton, Reizenstein, Rungta, Saladi, Schelten, Silva, Smith,
  Subramanian, Tan, Tang, Taylor, Williams, Kuan, Xu, Yan, Zarov, Zhang, Fan,
  Kambadur, Narang, Rodriguez, Stojnic, Edunov, and
  Scialom}]{touvron-et-al-2023-llama2}
Hugo Touvron, Louis Martin, Kevin Stone, Peter Albert, Amjad Almahairi, Yasmine
  Babaei, Nikolay Bashlykov, Soumya Batra, Prajjwal Bhargava, Shruti Bhosale,
  Dan Bikel, Lukas Blecher, Cristian Canton{-}Ferrer, Moya Chen, Guillem
  Cucurull, David Esiobu, Jude Fernandes, Jeremy Fu, Wenyin Fu, Brian Fuller,
  Cynthia Gao, Vedanuj Goswami, Naman Goyal, Anthony Hartshorn, Saghar
  Hosseini, Rui Hou, Hakan Inan, Marcin Kardas, Viktor Kerkez, Madian Khabsa,
  Isabel Kloumann, Artem Korenev, Punit~Singh Koura, Marie{-}Anne Lachaux,
  Thibaut Lavril, Jenya Lee, Diana Liskovich, Yinghai Lu, Yuning Mao, Xavier
  Martinet, Todor Mihaylov, Pushkar Mishra, Igor Molybog, Yixin Nie, Andrew
  Poulton, Jeremy Reizenstein, Rashi Rungta, Kalyan Saladi, Alan Schelten, Ruan
  Silva, Eric~Michael Smith, Ranjan Subramanian, Xiaoqing~Ellen Tan, Binh Tang,
  Ross Taylor, Adina Williams, Jian~Xiang Kuan, Puxin Xu, Zheng Yan, Iliyan
  Zarov, Yuchen Zhang, Angela Fan, Melanie Kambadur, Sharan Narang,
  Aur{\'{e}}lien Rodriguez, Robert Stojnic, Sergey Edunov, and Thomas Scialom.
  2023.
\newblock \href {https://doi.org/10.48550/ARXIV.2307.09288} {Llama 2: Open
  foundation and fine-tuned chat models}.
\newblock \emph{CoRR}, abs/2307.09288.

\bibitem[{von Werra et~al.(2020)von Werra, Belkada, Tunstall, Beeching, Thrush,
  Lambert, and Huang}]{vonwerra2022trl}
Leandro von Werra, Younes Belkada, Lewis Tunstall, Edward Beeching, Tristan
  Thrush, Nathan Lambert, and Shengyi Huang. 2020.
\newblock Trl: Transformer reinforcement learning.
\newblock \url{https://github.com/huggingface/trl}.

\bibitem[{Voorhees and Tice(2000)}]{voorhees-tice-2000-trec}
Ellen~M. Voorhees and Dawn~M. Tice. 2000.
\newblock \href
  {http://www.lrec-conf.org/proceedings/lrec2000/html/summary/26.htm} {The
  {TREC-8} question answering track}.
\newblock In \emph{Proceedings of the Second International Conference on
  Language Resources and Evaluation, {LREC} 2000, 31 May - June 2, 2000,
  Athens, Greece}. European Language Resources Association.

\bibitem[{Wang et~al.(2023)Wang, Cheng, Guo, Yue, Ding, Xu, Wang, Hu, Zhang,
  and Zhang}]{wang2023evaluating}
Cunxiang Wang, Sirui Cheng, Qipeng Guo, Yuanhao Yue, Bowen Ding, Zhikun Xu,
  Yidong Wang, Xiangkun Hu, Zheng Zhang, and Yue Zhang. 2023.
\newblock \href
  {https://proceedings.neurips.cc/paper_files/paper/2023/file/f323d594aa5d2c68154433a131c07959-Paper-Datasets_and_Benchmarks.pdf}
  {Evaluating open-qa evaluation}.
\newblock In \emph{Advances in Neural Information Processing Systems},
  volume~36, pages 77013--77042. Curran Associates, Inc.

\bibitem[{Wei et~al.(2022)Wei, Wang, Schuurmans, Bosma, Ichter, Xia, Chi, Le,
  and Zhou}]{wei-et-al-2022-cot}
Jason Wei, Xuezhi Wang, Dale Schuurmans, Maarten Bosma, Brian Ichter, Fei Xia,
  Ed~H. Chi, Quoc~V. Le, and Denny Zhou. 2022.
\newblock \href
  {http://papers.nips.cc/paper\_files/paper/2022/hash/9d5609613524ecf4f15af0f7b31abca4-Abstract-Conference.html}
  {Chain-of-thought prompting elicits reasoning in large language models}.
\newblock In \emph{Advances in Neural Information Processing Systems 35: Annual
  Conference on Neural Information Processing Systems 2022, NeurIPS 2022, New
  Orleans, LA, USA, November 28 - December 9, 2022}.

\bibitem[{Yang et~al.(2023)Yang, Chern, Qiu, Neubig, and
  Liu}]{yang-etal-2023-alignment}
Yuqing Yang, Ethan Chern, Xipeng Qiu, Graham Neubig, and Pengfei Liu. 2023.
\newblock \href {https://doi.org/10.48550/ARXIV.2312.07000} {Alignment for
  honesty}.
\newblock \emph{CoRR}, abs/2312.07000.

\bibitem[{Zhang et~al.(2023)Zhang, Chen, Xu, Cao, Chen, Cohn, and
  Fang}]{zhang-et-al-2023-survey-efficient}
Qin Zhang, Shangsi Chen, Dongkuan Xu, Qingqing Cao, Xiaojun Chen, Trevor Cohn,
  and Meng Fang. 2023.
\newblock \href {https://doi.org/10.18653/v1/2023.acl-long.808} {A survey for
  efficient open domain question answering}.
\newblock In \emph{Proceedings of the 61st Annual Meeting of the Association
  for Computational Linguistics (Volume 1: Long Papers)}, pages 14447--14465,
  Toronto, Canada. Association for Computational Linguistics.

\bibitem[{Zhang et~al.(2020)Zhang, Kishore, Wu, Weinberger, and
  Artzi}]{zhang-et-al-2020-bertscore}
Tianyi Zhang, Varsha Kishore, Felix Wu, Kilian~Q. Weinberger, and Yoav Artzi.
  2020.
\newblock \href {https://openreview.net/forum?id=SkeHuCVFDr} {{BERTScore}:
  Evaluating text generation with {BERT}}.
\newblock In \emph{8th International Conference on Learning Representations,
  {ICLR} 2020, Addis Ababa, Ethiopia, April 26-30, 2020}. OpenReview.net.

\bibitem[{Zhong et~al.(2023)Zhong, Ding, Liu, Du, and
  Tao}]{zhong-etal-2023-chatgpt}
Qihuang Zhong, Liang Ding, Juhua Liu, Bo~Du, and Dacheng Tao. 2023.
\newblock \href {https://doi.org/10.48550/ARXIV.2302.10198} {Can chatgpt
  understand too? {A} comparative study on chatgpt and fine-tuned {BERT}}.
\newblock \emph{CoRR}, abs/2302.10198.

\end{thebibliography}
\appendix

\section{Entailment Test Implementation}
\label{app:implementation}

\subsection{Detailed Settings}
\label{app:settings}

When using entailment to obtain the answer hierarchy in \S\ref{sec:hierarchy}, we use \texttt{gpt-3.5-turbo-1106}.
The gold answer-question and system answer-question pairs are converted to two declarative statements using the first prompt in Table~\ref{tab:entailment-prompts}.
The two examples in the first prompt are chosen from \textsc{Evouna}-NQ, and as the dataset size is large enough, we do not need to use a separate dataset for prompt engineering.
The two declarative statements are then used as the premise and hypothesis and vice versa in the second prompt in Table~\ref{tab:entailment-prompts} to obtain the entailment classification in two directions.
For all GPT-3.5 API calls, we set \texttt{seed$=$42, temperature$=$0.0} to ensure reproducible results, and set \texttt{max\_tokens$=$300}.

Here is a working example of the entailment test on a system answer generated by InstructGPT and a gold answer from NQ:

\begin{quote}
    \underline{Question:} where is fe best absorbed in the body\newline
    \underline{Gold answer:} in the duodenum \newline
    \underline{System answer:} Iron is best absorbed in the small intestine.\newline
    \newline
    \underline{Gold statement:} Fe is best absorbed in the body in the duodenum.\newline 
    \underline{System statement:} Iron is best absorbed in the small intestine.\newline
    \newline
    \underline{Entailment test:} Gold statement entails system statement, but not the other way around. Therefore, the system answer belongs to $A_{inf}-A_{sup}$. Meanwhile, human annotator judged the system answer as correct in \textsc{Evouna}.
\end{quote}

\subsection{Assessment of Reliability}
\label{app:reliability}

To assess the statistical reliability (consistency) of our method, we measure the agreement across different random seeds, and the potential impact on the overall performance.
We repeat the LLM-based steps on \textsc{Evouna}-NQ and \textsc{Evouna}-TQ subsets of size 2,000, each using random seeds 0,1,2,3 for GPT3.5 calls while keeping the rest of the settings controlled.

\paragraph{Reliability of question-answer to statement conversion.}
We calculate the consistency of generated statements from the same question-answer pairs across different seeds using BLEU and exact sentence matching, as shown in Table~\ref{tab:reliability-qa2d}. These results indicate that the generated statements are fairly consistent across different seeds with almost all statements being identical or very similar.

\begin{table}[h]
    \centering
    \begin{tabular}{lcc}
    \hline
    \textbf{Dataset} & \textbf{BLEU} & \textbf{Exact Match} \\ \hline
    NQ & 93.9 $\pm$ 1.6 & 86.7\% $\pm$ 2.1 \\
    TQ & 94.7 $\pm$ 0.1 & 83.6\% $\pm$ 0.2 \\ \hline
    \end{tabular}
    \caption{Reliability of question-answer to statement conversion, measured by average pairwise BLEU scores and percentages of exact matches across three runs.}
    \label{tab:reliability-qa2d}
\end{table}

\paragraph{Reliability of textual entailment test.}
We measure the agreement of textual entailment predictions across different pairs of seeds for the same golden-system answer pairs using Cohen's Kappa, as in Table~\ref{tab:reliability-entailment}. The results are interpreted as \emph{almost perfect} agreement according to \citeposs{landis-koch-1977-kappa} guideline.

\begin{table}[h]
    \centering
    \resizebox{\columnwidth}{!}{
    \begin{tabular}{lcccccc}
    \hline
    \textbf{Dataset} & \textbf{0 vs 1} & \textbf{0 vs 2} & \textbf{0 vs 3} & \textbf{1 vs 2} & \textbf{1 vs 3} & \textbf{2 vs 3}  \\ \hline
    NQ & 0.902 & 0.900 & 0.902 & 0.922 & 0.917 & 0.920 \\
    TQ & 0.873 & 0.882 & 0.870 & 0.872 & 0.870 & 0.865 \\ \hline
    \end{tabular}
    }
    \caption{Reliability of textual entailment test, measured by pairwise Cohen's Kappa across three runs.}
    \label{tab:reliability-entailment}
\end{table}

\paragraph{Reliability of hierarchy construction.} The result from textual entailment is used to categorize a system answer into one of the sets in the hierarchy (Table \ref{tab:nouva-nq-hierarchy} and \ref{tab:nouva-tq-hierarchy}). Again we measure the agreement of the categorization across different pairs of seeds using Cohen's Kappa.
The results in Table~\ref{tab:reliability-hierarchy} show even better agreement than the textual entailment test as multiple candidate golden answers are considered in this step.

\begin{table}[h]
    \centering
    \resizebox{\columnwidth}{!}{
    \begin{tabular}{lcccccc}
    \hline
    \textbf{Dataset} & \textbf{0 vs 1} & \textbf{0 vs 2} & \textbf{0 vs 3} & \textbf{1 vs 2} & \textbf{1 vs 3} & \textbf{2 vs 3}  \\ \hline
    NQ & 0.906 & 0.912 & 0.907 & 0.932 & 0.928 & 0.929 \\
    TQ & 0.921 & 0.925 & 0.924 & 0.927 & 0.917 & 0.917 \\ \hline
    \end{tabular}
    }
    \caption{Reliability of answer hierarchy construction, measured by pairwise Cohen's Kappa across three runs.}
    \label{tab:reliability-hierarchy}
\end{table}

\paragraph{Reliability of QA evaluation.}
We assess whether different seeds lead to different QA system evaluation results (Table~\ref{tab:eval-eval}) as reflected by the variance of F1 scores and accuracy. On the two subsets, different seeds have virtually no impact on the overall QA evaluation as seen in Table~\ref{tab:reliability-eval}.

\begin{table}[h]
    \centering
    \begin{tabular}{lcc}
    \hline
    \textbf{Dataset} & \textbf{F$_1$} & \textbf{Accuracy} \\ \hline
    NQ & 0.918 $\pm$ 0.002 & 0.876 $\pm$ 0.003\\
    TQ & 0.962 $\pm$ 0.001 & 0.934 $\pm$ 0.001 \\ \hline
    \end{tabular}
    \caption{Reliability of QA evaluation, measured by the variance of F1 scores and accuracy across three runs.}
    \label{tab:reliability-eval}
\end{table}

\subsection{Assessment of Validity}
\label{app:validity}
Converting a question-answer pair to a declarative statement (known as QA2D) is a well-established task.
\citet{demszky-etal-2018-qa2d} provided a dataset where the dev set has 10,344 question-answer pairs with a human-written declarative statement (test set unavailable).
We compare our 2-shot LLM generated statements with the human-written statements using BLEU and ROGUE (Table~\ref{tab:qa2d-eval}), and the generations are very similar to human-written statements and a fine-tuned T5 baseline\footnote{\href{https://huggingface.co/domenicrosati/QA2D-t5-base}{\texttt{domenicrosati/QA2D-t5-base}} on Huggingface.}.

\begin{table}[h]
    \centering
    \resizebox{\columnwidth}{!}{
    \begin{tabular}{lcccc}
    \hline
    \textbf{Model} & \textbf{BLEU} & \textbf{ROGUE-1} & \textbf{ROGUE-2} & \textbf{ROGUE-L} \\ \hline
    GPT-3.5 & 72.5 & 92.5 & 83.5 & 85.8 \\
    T5 & 72.7 & 90.1 & 82.4 & 85.8 \\ \hline
    \end{tabular}
    }
    \caption{\label{tab:qa2d-eval}
    Comparison of generated declarative statements with human-written statements on QA2D dataset.
    }
\end{table}

Our zero-shot prompt for textual entailment (Table~\ref{tab:entailment-prompts}) is adapted from \citet{qin-et-al-2023-chatgpt}.
They have tested the validity of this textual entailment test method on RTE and CB datasets and reported a high accuracy of 0.86 and 0.89 respectively. \citet{zhong-etal-2023-chatgpt} used a slightly different prompt for the same task and reported GPT-3.5 NLI accuracy to be higher than finetuned BERT-large and RoBERTa-large on both MNLI-m and RTE.

Finally, the validity of final QA evaluation is confirmed by the high correlation with human judgment (Table~\ref{tab:eval-eval}).

\section{Inter-set Order Validation}

\subsection{Hierarchy of Answer Sets}
\label{app:hierarchy}

The entailment test organizes system answers into a hierarchy of sets: in Table~\ref{tab:nouva-nq-hierarchy} and \ref{tab:nouva-tq-hierarchy}, the four rows corresponds to the four sets at different levels of the hierarchy: (1) $A_{sup}-A_{inf}$, (2) $A_{sup}\cap A_{inf}$, (3) $A_{inf}-A_{sup}$, and (4) $S-(A_{sup}\cup A_{inf})$. The size of the sets are shown in the \textit{Count} column.

\begin{table}[h]
    \centering
    \begin{tabular}{cccc}
    \hline
    \textbf{Rank} & \textbf{in} $A_{sup}$ & \textbf{in} $A_{inf}$ & \textbf{Count} \\ \hline
    (1) & Yes & No & 514  \\
    (2) & Yes & Yes  & 10,061  \\
    (3) & No & Yes & 1,000 \\
    (4) & No & No & 3,470\\ \hline
    \end{tabular}
    \caption{Distribution of system answers in different sets of the answer hierarchy for \textsc{Evouna}-NQ.}
    \label{tab:nouva-nq-hierarchy}
\end{table}

\begin{table}[h]
    \centering
    \begin{tabular}{cccc}
    \hline
    \textbf{Rank} & \textbf{in} $A_{sup}$ & \textbf{in} $A_{inf}$ & \textbf{Count} \\ \hline
    (1) & Yes & No & 168  \\
    (2) & Yes & Yes  & 7,890  \\
    (3) & No & Yes & 460 \\
    (4) & No & No & 1,172\\ \hline
    \end{tabular}
    \caption{Distribution of system answers in different sets of the answer hierarchy for \textsc{Evouna}-TQ.}
    \label{tab:nouva-tq-hierarchy}
\end{table}

We propose to treat the union (denoted as $\cup$) of $A_{inf}$ and $A_{sup}$ as the correct system answers.
We chose to include $A_{inf}-A_{sup}$ as we follow the human annotation guideline of \textsc{Evouna} that considers the lack of specificity in questions and accepts answers of all levels of specificity.
Meanwhile, alternative choices like excluding $A_{inf}-A_{sup}$ (denoted as $-$) have negligible impact on the discussion, as the $A_{inf}-A_{sup}$ sets have a small size for both NQ and TQ, as shown in Table~\ref{tab:nouva-nq-hierarchy} and \ref{tab:nouva-tq-hierarchy}. We report the evaluation results of excluding $A_{inf}-A_{sup}$ in Table~\ref{tab:eval-eval-excl}.

\begin{table*}[t]
    \centering
    \resizebox{0.87\textwidth}{!}{
    \begin{tabular}{cllllllllll}
    \hline
                  & \multicolumn{2}{c}{\textbf{DPR-FiD}}                & \multicolumn{2}{c}{\textbf{InstructGPT}}            & \multicolumn{2}{c}{\textbf{ChatGPT}}                & \multicolumn{2}{c}{\textbf{GPT-4}}                  & \multicolumn{2}{c}{\textbf{BingChat}}               \\
                  Evaluator & \multicolumn{1}{c}{$F_1$} & \multicolumn{1}{c}{Acc} & \multicolumn{1}{c}{$F_1$} & \multicolumn{1}{c}{Acc} & \multicolumn{1}{c}{$F_1$} & \multicolumn{1}{c}{Acc} & \multicolumn{1}{c}{$F_1$} & \multicolumn{1}{c}{Acc} & \multicolumn{1}{c}{$F_1$} & \multicolumn{1}{c}{Acc} \\ \hline
    Entailment ($\cup$)    & 94.8                      & 92.5                    & 92.7             & 90.2           & 92.6             & 88.9           & 93.8             & 90.1           & 92.6             & 88.1           \\
    Entailment ($-$)    & 95.1                      & 93.1                    & 92.5             & 90.5           & 91.6             & 88.0           & 93.6             & 90.1           & 92.3            & 87.6           \\\hline
    \multicolumn{11}{c}{on \textsc{Evouna}-NaturalQuestions}                                                                                                                                                                                                                                             \\ \hline
    Entailment ($\cup$)    &  96.8 & 94.7 & 96.6 & 94.2 & 96.6 & 94.2 & 97.4 & 95.3 & 95.9 & 92.9 \\
    Entailment ($-$)    & 96.2 & 93.9 & 95.0 & 92.3 & 96.5 & 94.2 & 96.7 & 94.2 & 94.8 & 90.9 \\\hline
    \multicolumn{11}{c}{on \textsc{Evouna}-TriviaQA}                                                                                                                                                                                                                                             \\ \hline
    \end{tabular}
    } %

    \caption{\label{tab:eval-eval-excl}
    When $A_{inf}-A_{sup}$ is excluded from the judged correct answers (denoted as $-$), the evaluation results of various systems do not change significantly compared to when $A_{inf}-A_{sup}$ is included (denoted as $\cup$).
    Our discussion in \S\ref{sec:hierarchy} is not affected by the choice of including $A_{inf}-A_{sup}$.
    }
\end{table*}

\subsection{Statistical Tests}
\label{app:interset}

As summarized in \S\ref{sec:hierarchy}, we conduct statistical tests to verify if the four sets do have a order.
We hypothesize that the higher the rank, the more likely the system answer is correct.
We use one-tailed Fisher's exact test do a pairwise comparison the distribution of human judgements in the four sets with DPR-FiD, a method with extractive nature that makes semantic understanding excessive and lexical matching sufficient, excluded.
The results are shown in Table~\ref{tab:fisher-nouva}.

\begin{table}[h]
    \centering
    \begin{tabular}{cccc}
    \hline
    \textbf{Dataset} & \textbf{Test} & \textbf{odds ratio} & \textbf{$p$} \\ \hline
    NQ & (1)>(2) & 1.35 & 0.008  \\
       & (2)>(3) & 2.59 & 2e-40  \\
       & (3)>(4) & 6.25 & 8e-108 \\\hline
    TQ & (1)>(2) & 0.17 & N/A \\
       & (2)>(3) & 5.22 & 3e-46 \\
       & (3)>(4) & 7.88 & 8e-54 \\\hline
    \end{tabular}
    \caption{Results of Fisher's exact test for the answer hierarchy in \textsc{Evouna}.}
    \label{tab:fisher-nouva}
\end{table}

\section{Baseline Method Details}
\subsection{Unsupervised Evaluators}
\label{app:eval-baseline}

\citet{wang2023evaluating} evaluated multiple unsupervised evaluators, including lexical match, BERTScore, and GPT-3.5, on both \textsc{Evouna}-NQ and \textsc{Evouna}-TQ.
We make the comparisons with the numbers reported in their paper and refer the readers to \citet{wang2023evaluating} for the detailed settings of those baseline evaluators.
They also explored four additional prompting methods for the GPT-3.5 evaluator: Ignoring Background Information, Giving Reasons, Chain-of-Thought, and In-Context Learning, with the exact prompts provided in their paper.
For each category in Table~\ref{tab:eval-eval}, we choose the best performing method among the four for comparison as in the \textit{GPT-3.5 (best prompting)}.
This represents the upper bound performance of their prompt engineering efforts that is only achievable if an oracle exists that knows the best prompt for each QA system.

For \textit{Entailment (small)}, we use the same prompt as in Table~\ref{tab:entailment-prompts} row 1, but with 4-bit quantized \texttt{Llama-2-7B-GPTQ}\footnote{\href{https://huggingface.co/TheBloke/Llama-2-7B-GPTQ}{\texttt{TheBloke/Llama-2-7B-GPTQ}} on Huggingface.} instead of GPT-3.5 as the model for question to statement conversion.
We use a finetuned \texttt{DeBERTa-v3-large}\footnote{\href{https://huggingface.co/cross-encoder/nli-deberta-v3-large}{\texttt{cross-encoder/nli-deberta-v3-large}} on Huggingface.} by \citet{reimers-gurevych-2019-sbert} as the NLI model.

\subsection{Learned Evaluators}
\label{app:eval-sft}
We perform a half-half partition of the \textsc{Evouna}-NQ dataset by question type to create a training set and a test set, where no question-answer pairs with the same question falls in the same split.
A \texttt{Llama-2-7b-chat-hf} model is finetuned on the training set by inserting the question, gold answer, system answer, and human judgment into the templates in Table~\ref{tab:sft-prompts}.
During inference, the same templates are used with human judgment left empty.
Finetuning is done with the Huggingface \texttt{PEFT} \cite{peft} and \texttt{TRL} \cite{vonwerra2022trl} libraries.
For CIV, two models with and without system answers as rationales are finetuned on the training set in the same fashion using templates in Table~\ref{tab:cvi-prompts}.

\section{Partial Mark Scoring}
\label{app:partial}
If the entailment test shows that the system answer is in $A_{inf}-A_{sup}$, we use GPT-3.5 and the prompt in Table~\ref{tab:scoring-prompts} row 1 to generate an explanation of what inference process is required to deduce the system answer from the gold answer (\textit{Inference}). The example system answer $a_2$ in Figure~\ref{fig:workflow} is in $A_{inf}-A_{sup}$, and the explanation generated is as follows:
\begin{quote}
    1. The TV show the Curse of Oak Island is filmed on Oak Island. (Given in S1)

    2. Oak Island is located in Nova Scotia, Canada. [[INFO]]

    3. Therefore, the TV show the Curse of Oak Island is filmed in Nova Scotia, Canada. (Combining steps 1 and 2)
\end{quote}

Given the inference process explanation, we manually design the following partial mark scoring heuristics:
\begin{enumerate}
    \item \textbf{CIA}: \texttt{-\#Step*10}\texttt{-\#INFO*3}\texttt{-\#ASSUMPTION*5}
    \item \textbf{C}: \texttt{-\#Step*10}
    \item \textbf{IA}: \texttt{-\#INFO*3}\texttt{-\#ASSUMPTION*5}
\end{enumerate}

As an alternative, we use GPT-3.5 to score the difficulty of the inference process in the 5-point scale by providing the prompt in Table~\ref{tab:scoring-prompts} row 2 as an additional message after the explanation step (\textit{Inference+LLM Score}). The \textit{LLM Score} baseline skips the explanation step and directly use GPT-3.5 to provide a 5-point-scale score using the prompt in Table~\ref{tab:scoring-prompts} row 3.

The three manually designed scoring scheme are not significantly different from each other or from the automated \textit{Inference+LLM Score} as shown in Table~\ref{tab:partial-full}.

\begin{table}[h]
    \centering
    \begin{tabular}{lr}
    \hline
    \textbf{Method}                & \textbf{AUC} \\ \hline
    Inference + LLM Score & 0.9119  \\
    Inference + CIA    & 0.9120  \\
    Inference + IA    & 0.9118  \\
    Inference + C    & 0.9118  \\
    LLM Score             & 0.8827  \\
    $F_1$ Score           & 0.7770  \\ \hline
    \end{tabular}
    \caption{Area under the receiver operating characteristic curve (AUROC) in predicting human judgements on NQ system answers for more scoring schemes.}
    \label{tab:partial-full}
\end{table}

\begin{table*}[htbp]
    \centering
    \begin{tabularx}{\textwidth}{p{3cm} | X}
        \hline
        \textbf{Description}     & \textbf{Prompt} \\ \hline
        Convert question-answer pair 
        to a declarative statement  & Convert a question answer pair to a declarative statement, following these two examples:\newline
        Q: where is the tv show the curse of oak island filmed\newline
        A: Oak Island\newline
        S: The TV show the Curse of Oak Island is filmed on Oak Island.\newline
        \newline
        Q: who wrote the first declaration of human rights\newline
        A: Cyrus\newline
        S: Cyrus wrote the first declaration of human rights\newline
        \newline
        Do not provide explanations. Provide the statement only. Follow the above examples and convert this pair:\newline
        Q: \{question\}\newline
        A: \{answer\}\newline
        S:        \\ \hline
        Entailment test & Please identify whether the premise entails or contradicts the hypothesis in the following premise and hypothesis. The answer should be exact “entailment”, “contradiction”, or “neutral”. Provide only the answer from the three options. Do not provide explanations.\newline
        \newline
        Premise: \{premise\}\newline
        Hypothesis: \{hypothesis\}\newline
        \newline
        Is it entailment, contradiction, or neutral? \\ \hline
    \end{tabularx}
    \caption{
        \label{tab:entailment-prompts}
        Prompts for the entailment test. The second prompt adapted from \citet{qin-et-al-2023-chatgpt}.
    }
\end{table*}

\begin{table*}[htbp]
    \centering
    \begin{tabularx}{\textwidth}{p{3cm} | X}
        \hline
        \textbf{Description}     & \textbf{Prompt} \\ \hline
        Template for finetuning Llama-2 & <s> [INST] Here is a question, a set of golden answers (split with /), an AI-generated answer.\newline
        Can you judge whether the AI-generated answer is correct according to the question and golden answers, simply answer Yes or No.\newline
        \newline 
        Question: \{question\}\newline
        Golden answers: \{golden answer\}\newline
        AI answer: \{system\}\newline
        [/INST] \{system answer\} </s>\\\hline
        Template for finetuning Llama-2 with NLI as a feature & <s> [INST] Here is a question, a set of golden answers (split with /), an AI-generated answer.\newline
        Can you judge whether the AI-generated answer is correct according to the question and golden answers, simply answer Yes or No.\newline
        \newline 
        Question: \{question\}\newline
        Golden answers: \{golden answer\}\newline
        AI answer: \{system\}\newline
        Can golden answers be inferred from AI answer: \{yes or no\}\newline
        Can AI answer be inferred from golden answers: \{yes or no\}\newline
        [/INST] \{system answer\} </s> \\
         \hline
    \end{tabularx}
    \caption{
        \label{tab:sft-prompts}
        Prompts for finetuned Llama-2-7B evaluators.
    }
\end{table*}

\begin{table*}[htbp]
    \centering
    \begin{tabularx}{\textwidth}{p{5cm} | X}
        \hline
        \textbf{Description}     & \textbf{Prompt} \\ \hline
        Template for training the QA model with system answer as the rationale& <s> [INST] Given the fact: \{system answer\},\newline
        answer this question: \{question\}\newline
        [/INST] \{golden answer\} </s>\\\hline
        Template for training the QA model without rationales & <s> [INST] Answer this question: \{question\}\newline
        [/INST] \{golden answer\} </s>\\\hline
    \end{tabularx}
    \caption{
        \label{tab:cvi-prompts}
        Prompts for training the QA model with and without system answers as rationales (for CVI) by finetuning Llama-2.
    }
\end{table*}

\begin{table*}[htbp]
    \centering
    \begin{tabularx}{\textwidth}{p{3cm} | X}
        \hline
        \textbf{Description}     & \textbf{Prompt} \\ \hline
        \textit{Inference}: Explain the inference process  & We have two statements S1 (the premise) and S2 (the hypothesis). S1 entails S2.\newline
        \newline
        S1: \{s1\}\newline
        S2: \{s2\}\newline
        Now, list the reasoning process step by step to show how S2 can be deduced from S1.\newline
        List the steps as numbered statements, starting from 1.\newline
        If a step involves information not mentioned in S1 and S2, append {[}{[}INFO{]}{]} after the step.\newline
        If an assumption must be made to deduce S2 from S1, append {[}{[}ASSUMPTION{]}{]} after the step.\newline
        Provide the reasoning steps only. Do not include any other information.\\ \hline
        \textit{Inference + LLM Score}: Rate the inference difficulty based on the explanation & Based on the reasoning steps, rate how hard it is to deduce S2 from S1.\newline
        1: Very easy\newline
        2: Easy\newline
        3: Neither easy nor hard\newline
        4: Hard\newline
        5: Very hard\newline
        Consider how many assumptions are needed, how much information is needed, and how much reasoning is needed.\newline
        Return a number from 1 to 5 only. Do not include any other information.\\ \hline
        \textit{LLM Score}: Directly use LLM to provide a score of answer closeness & Here is a question, a set of golden answers (split with /), an AI-generated answer.
        Can you judge whether the AI-generated answer is correct according to the question and golden answers? Simply give a score from 1 to 5.\newline
        1: The AI-generated answer is completely wrong.\newline
        2: The AI-generated answer is mostly wrong.\newline
        3: The AI-generated answer is neither wrong nor right.\newline
        4: The AI-generated answer is mostly right.\newline
        5: The AI-generated answer is completely right.\newline
        \newline
        Question: \{question\}\newline
        Golden answers: \{golden answer\}\newline
        AI answer: \{system answer\}\\ \hline
    \end{tabularx}
    \caption{
        \label{tab:scoring-prompts}
        Prompts for generating the inference explanation and scoring the inference difficulty.
    }
\end{table*}

\end{document}